\documentclass[a4paper]{article}
\usepackage{balance}
\usepackage{INTERSPEECH2019}
\usepackage{xcolor}
\usepackage{hyperref}

\title{Building a mixed-lingual neural TTS system with only monolingual data}
\name{Liumeng Xue\thanks{Work was done when Liumeng Xue was an intern at JD.com. }$^{1}$, Wei Song$^2$, Guanghui Xu$^2$,  Lei Xie$^{1}$, Zhizheng Wu$^2$}

\address{
  $^1$Shaanxi Provincial Key Lab of Speech and Image Information Processing,\\  
  School of Computer Science, Northwestern Polytechnical University, Xi'an, China \\ 
  $^2$JD.com}
\email{\{lmxue, lxie\}@nwpu-aslp.org, \{songwei11, xuguanghui, zhizheng.wu\}@jd.com}

\begin{document}

\maketitle

\begin{abstract}
When deploying a Chinese neural Text-to-Speech (TTS) system, one of the challenges is to synthesize Chinese utterances with English phrases or words embedded. This paper looks into the problem in the encoder-decoder framework when only monolingual data from a target speaker is available. Specifically, we view the problem from two aspects: speaker consistency within an utterance and naturalness. We start the investigation with an average voice model which is built from multi-speaker monolingual data, i.e., Mandarin and English data. On the basis of that, we look into speaker embedding for speaker consistency within an utterance and phoneme embedding for naturalness and intelligibility, and study the choice of data for model training. We report the findings and discuss the challenges to build a mixed-lingual TTS system with only monolingual data.

\end{abstract}
\vspace{-3pt}
\section{Introduction}
When deploying a non-English Text-to-Speech (TTS) system, it is very common that we have to address the mixed-lingual problem. A mixed-lingual TTS system is expected to synthesize utterances with embedded phrases or words from a different language. A straightforward way to build a mixed-lingual TTS system is to use a bilingual speech database recorded by a bilingual speaker. However, it's very hard to find a speaker with excellent multiple language skills and consistent articulation across different languages, and it is not flexible to use pre-recorded data which only has monolingual data. In this work, we look into the mixed-lingual TTS in Mandarin Chinese context with English phrases or words embedded.
\vspace{-4pt}
\subsection{Related work}\vspace{-3pt}
%The first sentence would be deleted.

A mixed-lingual TTS system is expected to generate high-quality speech and be perceived as spoken by the same speaker even  when switching languages in mixed-lingual utterances. Several studies have been conducted to assess the mixed-lingual problem. In~\cite{ traber1999multilingual}, Traber \textit{et al.} builds a mixed-lingual TTS system using a bilingual speech database recorded by a bilingual speaker. In~\cite{latorre2005polyglot}, HMM states are shared across languages and speech data from multiple languages are used. In~\cite{liang2007hmm,qian2008hmm,qian2009cross}, Mandarin and English context-dependent HMM states are shared, and the mapping is learned from a bilingual dataset recorded by a bilingual speaker.  In~\cite{he2012turning}, He \textit{et al.} proposes an approach to convert a monolingual TTS into multilingual by employing a bi-linear frequency warping function, and taking into account of cross-language F0 variations and equalizing speaking rate difference between source speaker and reference speaker. Besides, speaker adaptation and voice conversion are also effective ways to mixed-lingual speech synthesis using a set of monolingual or multilingual speech databases. In~\cite{ramani2013voice}, Ramani \textit{et al.} creates a polyglot corpus using voice conversion on a set of multilingual speech databases including Tamil, Telugu, Malayalam, and Hindi. The HMM-based polyglot TTS built with the polyglot database can synthesize mixed-lingual speech for four languages in target speaker voice. In~\cite{sitaram2016speech,sitaram2016experiments}, Sitaram \textit{et al.} presents a code-mixed TTS framework, in which the languages are not written in their native script but borrow the script of the other language. Then, the mapping between the phonemes of both languages is used to synthesize the text using a TTS system trained on a single language. In~\cite{chandu2017speech}, Chandu \textit{et al.} further extends their code-mixed TTS to a bilingual system using two monolingual speech datasets and a combined phone set for speech synthesis of mixed-language navigation instructions. For deep neural network based speech synthesis, a cross-lingual TTS is built using Kullback-Leibler divergence~\cite{xie2016kl}.

%{\color{blue}cross-lingual or mixed-lingual papers of Frank song} \cite{xie2016kl}
%The  above  mixed-lingual  systems  are  mostly  based  on HMM. For deep neural network (DNN) , a cross-lingual TTS is built using the Kullback-Leibler divergence (KLD) approach \cite{xie2016kl}. In this method, a speaker independent DNN ASR model is trained to equalize the speaker difference in different languages and KLD is used to measure the phonetic distortion between two given acoustic segments at senone or frame level probabilistically.

Recently, encoder-decoder framework has been successfully applied to TTS system. In ~\cite{li2018bytes}, Li \textit{et al.} presents two end-to-end models: Audio-to-Byte (A2B) and Byte-to-Audio (B2A), for multilingual speech recognition and synthesis, modeling text using a sequence of Unicode bytes, specifically, the UTF-8 variable length byte sequence for each character. The B2A model is able to synthesize code-switching text and the speech is fluent, but the speaker voice is changed for different language.
\vspace{-4pt}
\subsection{The contribution}\vspace{-3pt}
We conduct investigations based on the encoder-decoder framework, which is proven to generate speech with better prosody. In this study, we attempt to answer the following questions:
\begin{itemize}
\item Can the encoder-decoder model learn meaningful phonetic representations in encoder part? Does the encoder interpret Mandarin and English phonemes differently?
\item What is the impact of speaker embedding on speaker consistency within mixed-lingual utterances? 
\item Can phonetic information, i.e., phoneme embedding, be used in attention alignment and context vector and as a result improve naturalness and speaker consistency when switching languages in an utterance?
\item Is monolingual data enough to build a mixed-lingual TTS system?
\end{itemize}

To answer these questions, we conduct analysis on phoneme embeddings to understand what the encoder-decoder model is learning, and present phoneme-informed attention for the encoder-decoder model. Besides, we compare speaker embedding at different positions to analyze its effects on speaker consistency among mixed-lingual utterances. We also study the way of model training in terms of the choice of training data. 

%To answer these questions, we conduct analysis on phoneme embedding to understand what the encoder-decoder model is learning, and also study the way of model training in terms of the choice of training data. We also present the phoneme-informed attention for the encoder-decoder model.

%1, retrain speaker embedding or retrain AVM with target speaker's data
%
%2, analysis of phoneme embedding and encoder output, and study the relationship between Chinese and English phonemes
%
%3, study the use of residual encoder for intelligibility
%
%4, compare different training data strategy for model training
\vspace{-6pt}
\section{Mixed-lingual neural TTS system}
Even though there are two languages in one utterance, a mixed-lingual TTS system is expected to produce synthesized utterances that sound like one speaker. Hence, speaker consistency within an utterance, intelligibility and naturalness are all important factors for the mixed-lingual TTS.

It will be very challenging if not impossible to learn bilingual phonetic coverage when only monolingual data is available. We start our investigations from building an average voice model (AVM) with multi-speaker monolingual dataset.  \iffalse , specifically, the data from 35 monolingual Mandarin Chinese speakers and 35 monolingual English speakers.\fi We note that the Chinese corpus does not have English words and the English corpus does not have Chinese pronunciations. To control speaker consistency, speaker embedding is investigated, and phoneme embedding is also studied to better understand how the encoder-decoder model learns phonetic information.

\vspace{-4pt}
\subsection{Multi-speaker voice modeling}\vspace{-3pt}
There are two ways to build an AVM from multi-speaker data. One way is to mix all the data together and treat the data as from a single speaker. Retrain and adaptation can be performed on top of that. The other way is to assign each speaker a label (e.g., speaker code), and use the label to distinguish data from different speakers \iffalse in the model training process\fi. In this work, the second way is used and speaker embedding is applied. We used the first way to analyze phoneme embedding.
%and how the encoder learn phonetic representations.
%we use speaker embeddings to distinguish speakers, but when we analyze phoneme embeddings or phonetic representations from encoder, the former AVM training method is used. Experimental comparison is also performed for the two methods. 
% For the multi-speaker voice model, we first build an average voice model, in which we ignore the speaker characteristics, to investigate if bilingual phonetic information is learned using monolingual Mandarin Chinese data and monolingual English data. Then, we take each speaker voice into consideration, and build a multi-speaker voice model with speaker embedding to investigate if speaker embedding affects the speaker consistency within mixed-lingual utterances.

\vspace{-6pt}
\subsection{Speaker embedding}\vspace{-3pt}
We use speaker embedding to help the AVM training with multi-speaker monolingual data. The speaker embeddings are assumed to construct a speaker space. There are various approaches proposed for modeling the speaker space~\cite{Arik2018Neural,Nachmani2018Fitting,Ye2018Transfer}. Speaker embedding has been extensively used in multi-speaker speech synthesis to generate the speech of the specific speaker~\cite{Arik2018Neural,Nachmani2018Fitting,Taigman2017VoiceLoop,gibiansky2017deep,ping2018deep}. It has been proved that speaker embedding is an effective way to model a speaker space~\cite{li2017deep}. In general, speaker embedding can be concatenated with encoder output~\cite{Ye2018Transfer,Skerry2018Towards} or fed into decoder as an extra input~\cite{gibiansky2017deep}. In this paper, we use a speaker look-up table to store speaker embeddings, which is trained jointly with the encoder-decoder network. The speaker embeddings are utilized to condition speech synthesis to control speaker voice in both training and inference.  

To investigate how to place speaker embedding in the encoder-decoder architecture for better speaker consistency, we consider two different positions: 1) concatenating speaker embedding with the encoder output (SE-ENC) and 2) concatenating speaker embedding with the decoder input (SE-DEC) , i.e., with the LSTM input after prenet. To get the target speaker characteristic, we use two approaches: 1) excluding the target speaker data in the AVM but using it to retrain the decoder, 2) including the target speaker data in the AVM to learn speaker embeddings simultaneously.

\begin{figure}[t]
\setlength{\abovecaptionskip}{0.cm}
\setlength{\belowcaptionskip}{-0.cm}
  \centering
  \includegraphics[scale=0.48]{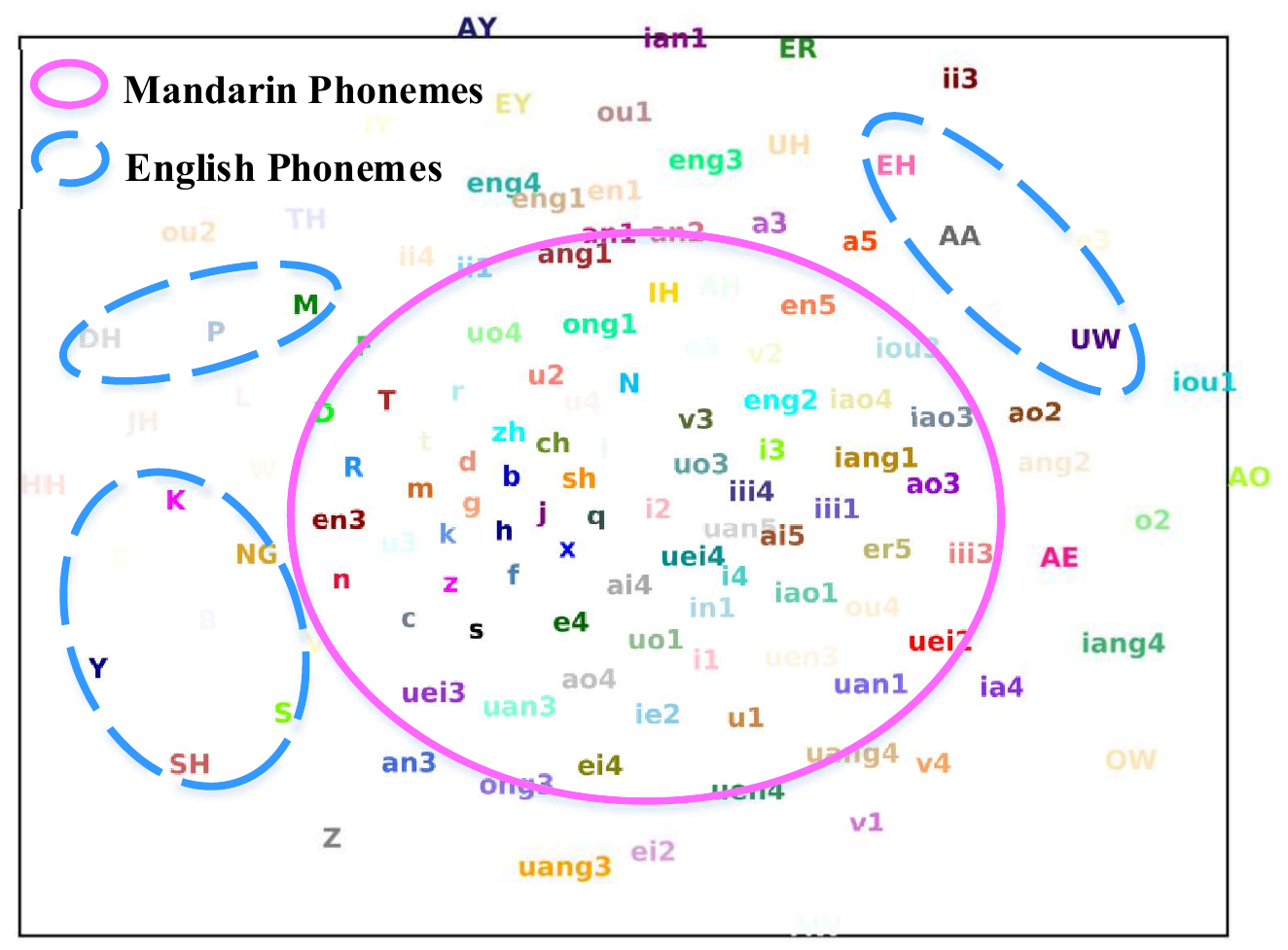}
 % \caption{Clustering result of phoneme embeddings.}
    \caption{Phoneme embeddings visualization using t-SNE.}
  \label{fig:1_1_cluster_phn_ori_nosil}\vspace{-14pt}
\end{figure}

\begin{figure}[t]
\setlength{\abovecaptionskip}{0.cm}
\setlength{\belowcaptionskip}{-0.cm}
  \centering
 \includegraphics[scale=0.48]{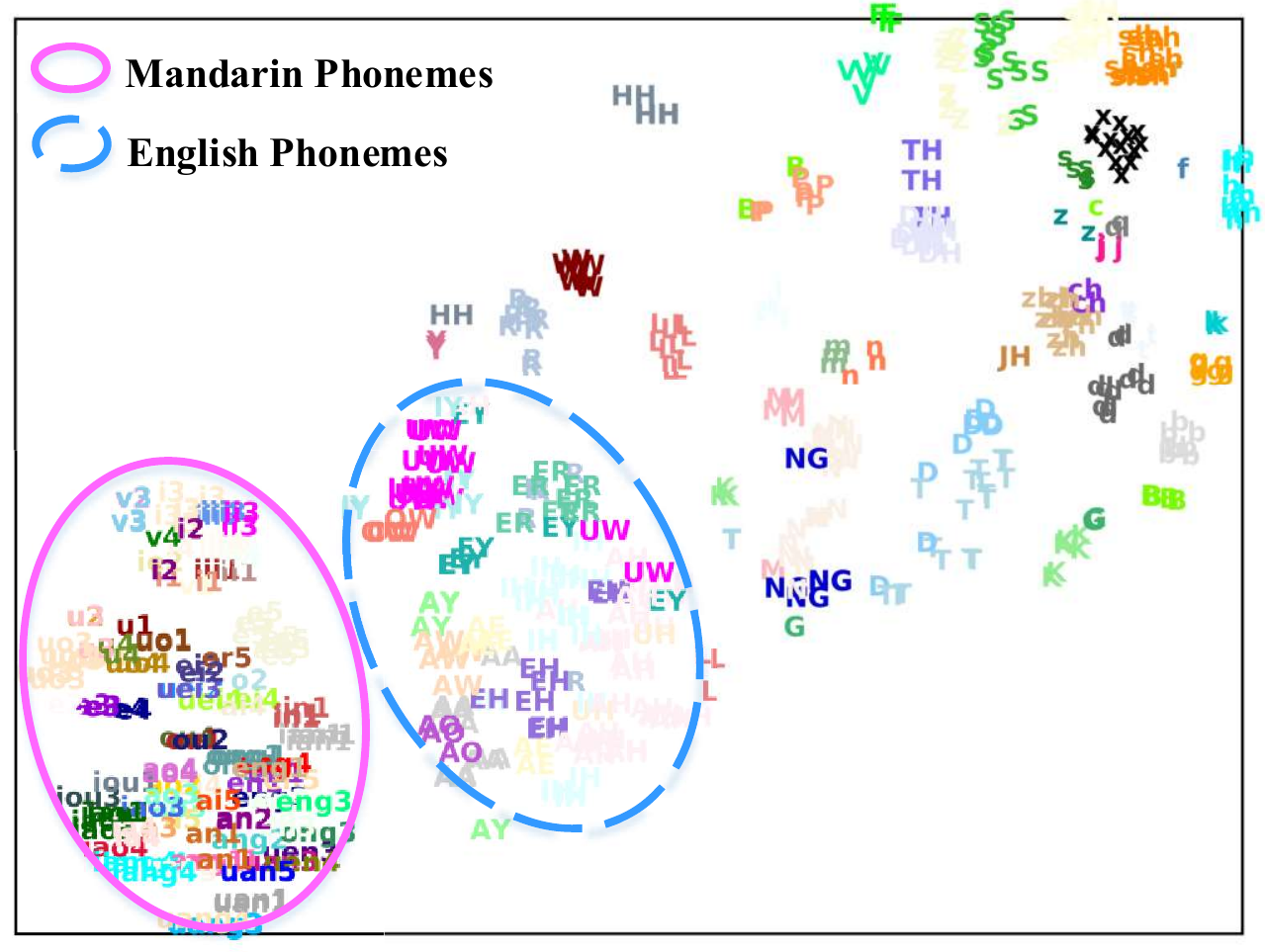}
 % \caption{Clustering result of encoder outputs.}
  \caption{Encoder outputs visualization using t-SNE.}
  \label{fig:1_2_clustering_enc_ori_nosil}\vspace{-17pt}
\end{figure}

\vspace{-5pt}
\subsection{Phoneme embedding}\vspace{-3pt}
Phonetic coverage and the relationship between English and Mandarin phonemes are important to the naturalness and intelligibility for a mixed-lingual TTS system. To this end, we analyze the phoneme embeddings and encoder outputs to understand how the encoder-decoder model learns phonetic representations \iffalse and how to interpret encoder outputs\fi. Phoneme embeddings are vectorized presentations of discrete phonemes, while encoder outputs are derived from phoneme embeddings, which also have a direct impact on the decoder via attention modeling and the resultant context vector.

Figure~\ref{fig:1_1_cluster_phn_ori_nosil} and Figure~\ref{fig:1_2_clustering_enc_ori_nosil} present the t-SNE~\cite{maaten2008visualizing} visualization of phoneme embeddings and encoder outputs, respectively. From the phoneme embedding representations, the English and Mandarin phonemes are separated in some sense but do not have a clear boundary like that in the visualization of encoder outputs.  In the visualization of encoder outputs, however, the clustering changes a bit. Mandarin and English phonemes are grouped into two separate clusters. It implies the properties of phoneme embeddings have been changed a bit after several layers of transformations. We suspect that the encoder output is affected more by the audio information which is passed down through back-propagation. We also argue that if the attention alignment is not accurate, and it may also introduce errors into encoder through back-propagation. Hence, it might be useful to have phonetic information when computing the attention alignment or context vector.

% In the phoneme embedding visualization, for Mandarin phonemes, phonemes with similar sound are clustered together, for example \textit{zh} and \textit{sh}, \textit{ao} and \textit{ou}. Regarding the relation between Mandarin and English phonemes, phonemes with similar sound also have a short distance in the visualization, for example \textit{A*} and \textit{a*} for English and Mandarin phonemes respectively.
% }

\vspace{-4pt}
\subsection{Phoneme-informed attention}
\begin{figure}[t]
  \centering
  \includegraphics[width=0.48\textwidth]{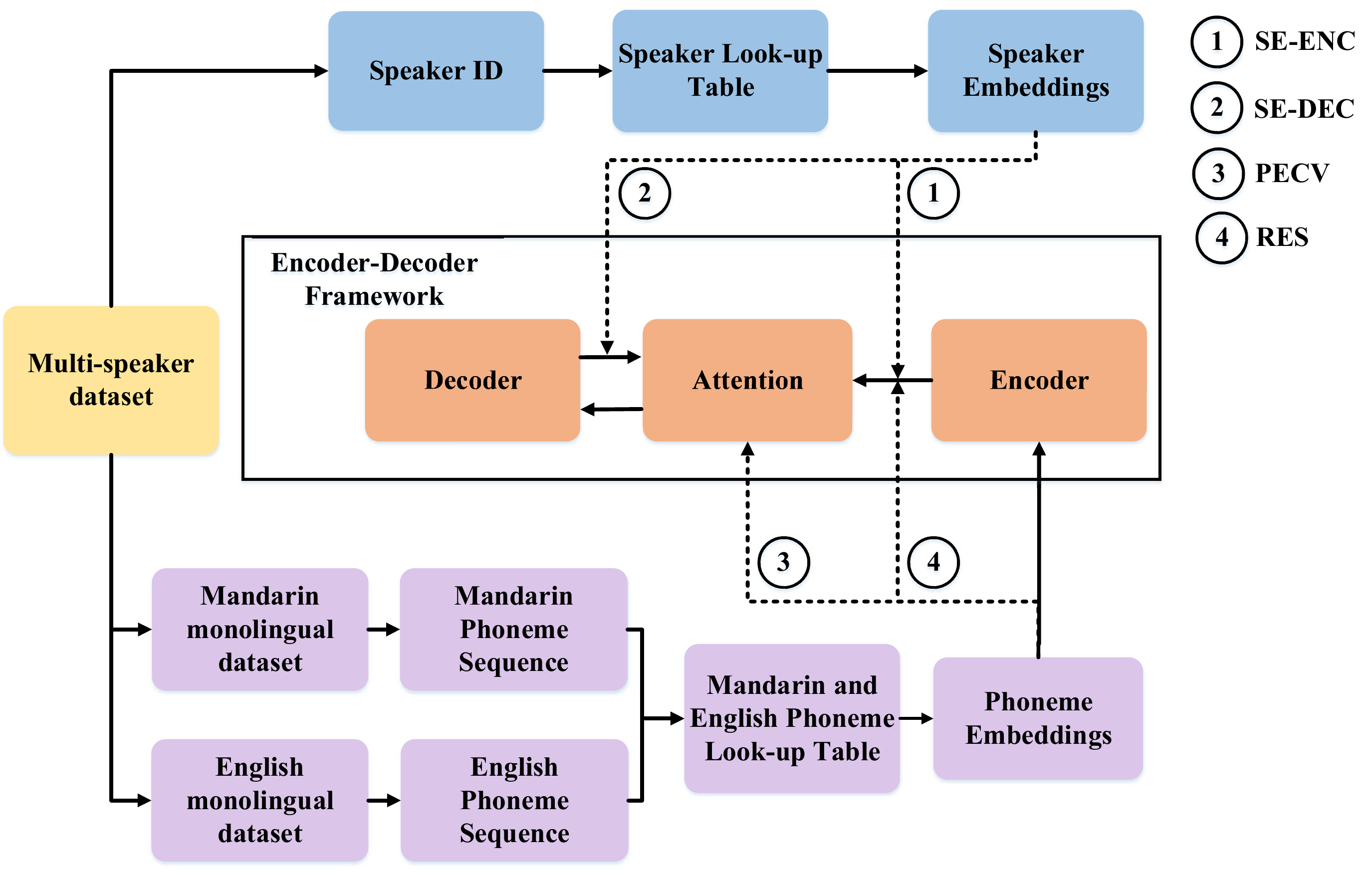}
  \caption{The architecture investigated in this study.}
  \label{fig:methods_list}\vspace{-4pt}
\end{figure}

\vspace{-3pt}
We investigate the impact of phoneme-informed attention from two directions. One is to calculate an additional phoneme embedding context vector (PECV) by applying attention weights to phoneme embeddings and concatenate it with the attention context vector. {\iffalse}which is fed to the decoder. As phoneme embeddings and encoder outputs have a one-to-one direct mapping, attention weights based on encoder outputs can be applied directly to phoneme embeddings for PECV. However, phoneme embeddings cannot have an impact to the attention alignment. {\fi}The other is to use a residual encoder (RES) architecture by adding the phoneme embeddings to encoder outputs directly. An illustration of the architecture investigated in this study can be found in Figure~\ref{fig:methods_list}. More details about the phoneme-informed attention are described in Sections 2.4.1 and 2.4.2.
%More details about the phoneme-informed attention can be found in the samples webpage\footnote{Samples can be found at \url{https://angelkeepmoving.github.io/mixed-lingual-tts/index.html} }.

\vspace{-4pt}
\subsubsection{Phoneme embedding context vector}\vspace{-3pt}
The phoneme embedding context vector is computed just like the attention context vector. In the attention mechanism~\cite{bahdanau2014neural}, a context vector $c_{i}$ of i-th decoder output step depends on the encoder outputs ($h_{1}$, $\cdots$ , $h_{T}$). The context vector is computed as a weighted sum of the encoder outputs $h_{j}$:
\begin{equation}
    c_{i} = \sum_{j=1}^{T}\alpha_{ij}h_{j}
\end{equation}
The weights $\alpha_{ij}$ of each $h_{j}$ is calculated as: 
\begin{equation}
    \alpha_{ij} = \frac{exp(e_{ij})}{\sum_{j=1}^{T}exp(e_{ij})}
\end{equation}
where 
\begin{equation}
   e_{ij} = a(s_{i-1}, h_{j})
\end{equation}
is the score computed on the decoder hidden state $s_{i-1}$ and the j-th representation $h_{j}$ of the encoder outputs. 

%In our work, the additional context vector $c'_{i}$ at i-th deocder output step is computed based on phoneme embeddings ($s_{1}$, $\cdots$ , $s_{T}$) and shares the same weights $\alpha_{ij}$ with the attention context vector $c_{i}$:
In our work, the additional context vector $c'_{i}$ at i-th deocder output step is computed based on phoneme embeddings ($p_{1}$, $\cdots$ , $p_{T}$) and shares the same weights $\alpha_{ij}$ with the attention context vector $c_{i}$:
\begin{equation}
%    c'_{i} = \sum_{j=1}^{T}\alpha_{ij}s_{j}
    c'_{i} = \sum_{j=1}^{T}\alpha_{ij}p_{j}
\end{equation}

Then, we concatenate the attention context vector and the phoneme embedding context vector as the new attention context vector $C_{i}$:
\begin{equation}
  C_{i} = [c_{i}; c'_{i}]
\end{equation}

Of course, dimension reduction is performed to the context vector before passing to the following network.

\vspace{-4pt}
\subsubsection{Residual encoder}
\begin{figure}[t]
   \centering
   \includegraphics[scale=0.25]{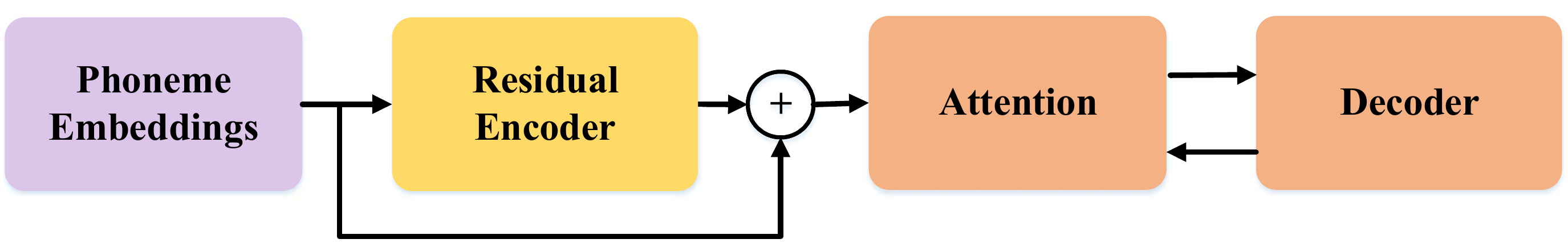}
   \caption{The structure of residual encoder.}
   \label{fig:residual_encoder}
 \end{figure}

\vspace{-3pt}
The structure of residual encoder is shown in Figure~\ref{fig:residual_encoder}, in which phoneme embeddings are added to encoder outputs to improve encoder representation. It should be noted that phoneme embeddings have the same dimensions as encoder outputs. 

The phoneme embedding context vector impact on the context vector only. While the residual encoder affects the whole attention calculation process, including scores, attention weights and context vector. We expected it can improve the overall performances of attention and resultant synthesized speech.

%{\color{red} In journal paper, we should have visualization to compare attention alignment with and without the residual encoder.}

\vspace{-3pt}
\section{Experimental Results and Analysis}\vspace{-1pt}
\subsection{Experimental setup}\vspace{-3pt}
In this paper, the goal is to study how to build a female Mandarin-English mixed voice using only Mandarin data. We limite the number of training utterances from the target speaker to 500. As the target speaker is a female, only female datasets are utilized to reduce the effect of gender factor. Hence, as we described earlier, to build an AVM, we use an internal Mandarin monolingual dataset from 35 female speakers and an English monolingual dataset from 35 female speakers with American accents, which is a subset of the public available dataset VCTK \cite{veaux2016superseded}. Each Mandarin monolingual speaker has around 500 utterances, in total of 17,197 utterances, which is approximately 17 hours of audio. Each English monolingual speaker has varied number of utterances from 200 to 500, in total of 14,464  utterances. English utterances are shorter than Mandarin utterances in duration, so the English dataset is about 8 hours of audio.

All audios are down-sampled to 24kHz. The beginning silence are all trimmed, and the ending silence are trimmed to a fixed length. 80-dimensional mel-spectrograms and 1024-dimensional linear spectrograms are extracted from audios as the model target output. Phoneme sequences are fed to the model as input to predict spectrograms. In this paper, experiments are performed based on the encoder-decoder neural TTS system \cite{shen2018natural}. Since our work focuses on generating mixed-lingual speech with satisfied intelligibility and a consistent voice, we use the Griffin-Lim \cite{griffin1984signal} algorithm to synthesize waveform from the predicted linear spectrograms like Tacotron-1 \cite{wang2017tacotron}, instead of using a WaveNet vocoder like Tacotron-2 \cite{shen2018natural}.

 \begin{figure}[t]

   \centering
   \includegraphics[scale=0.4]{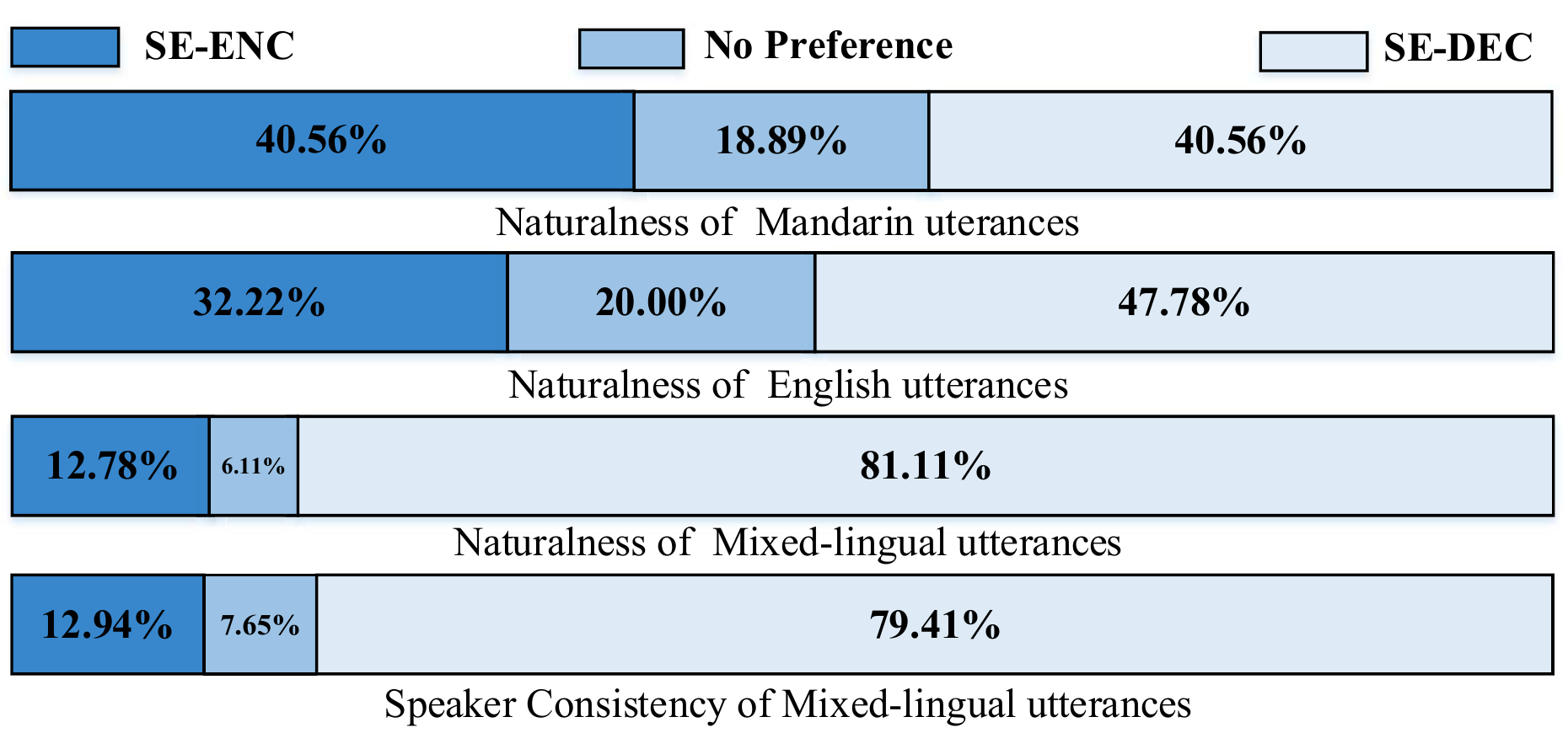}
   \caption{AB preference results of SE-ENC and SE-DEC.}
   \label{fig:ab_test_lang_enc_dec}\vspace{-4pt}
 \end{figure}
 
 \begin{figure}[t]
   \centering
   \includegraphics[scale=0.4]{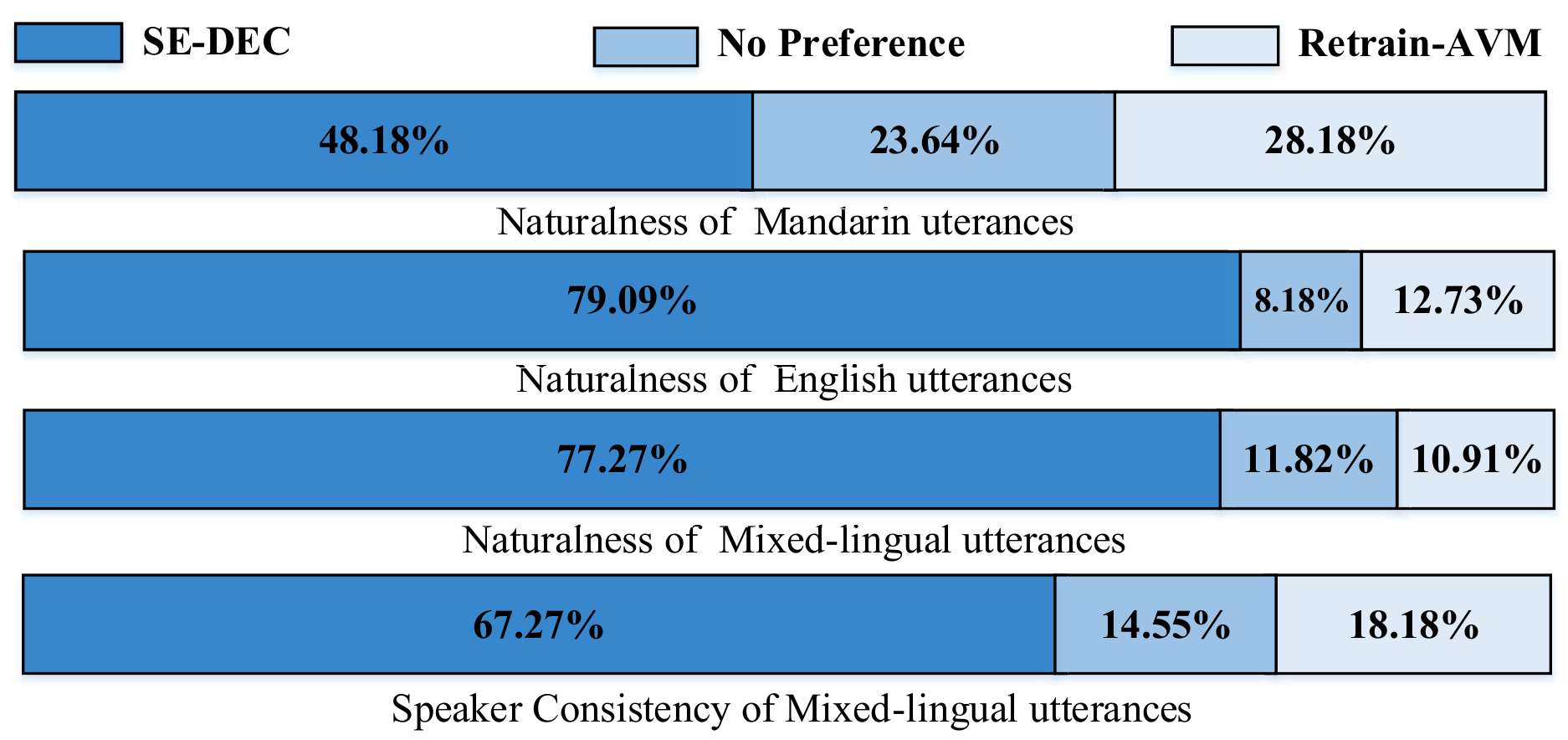}
   \caption{AB preference results of SE-DEC and Retrain-AVM.}
   \label{fig:ab_test_lang_retrain_dec}\vspace{-4pt}
 \end{figure}
 
  \begin{figure}[t]
   \centering
   \includegraphics[scale=0.4]{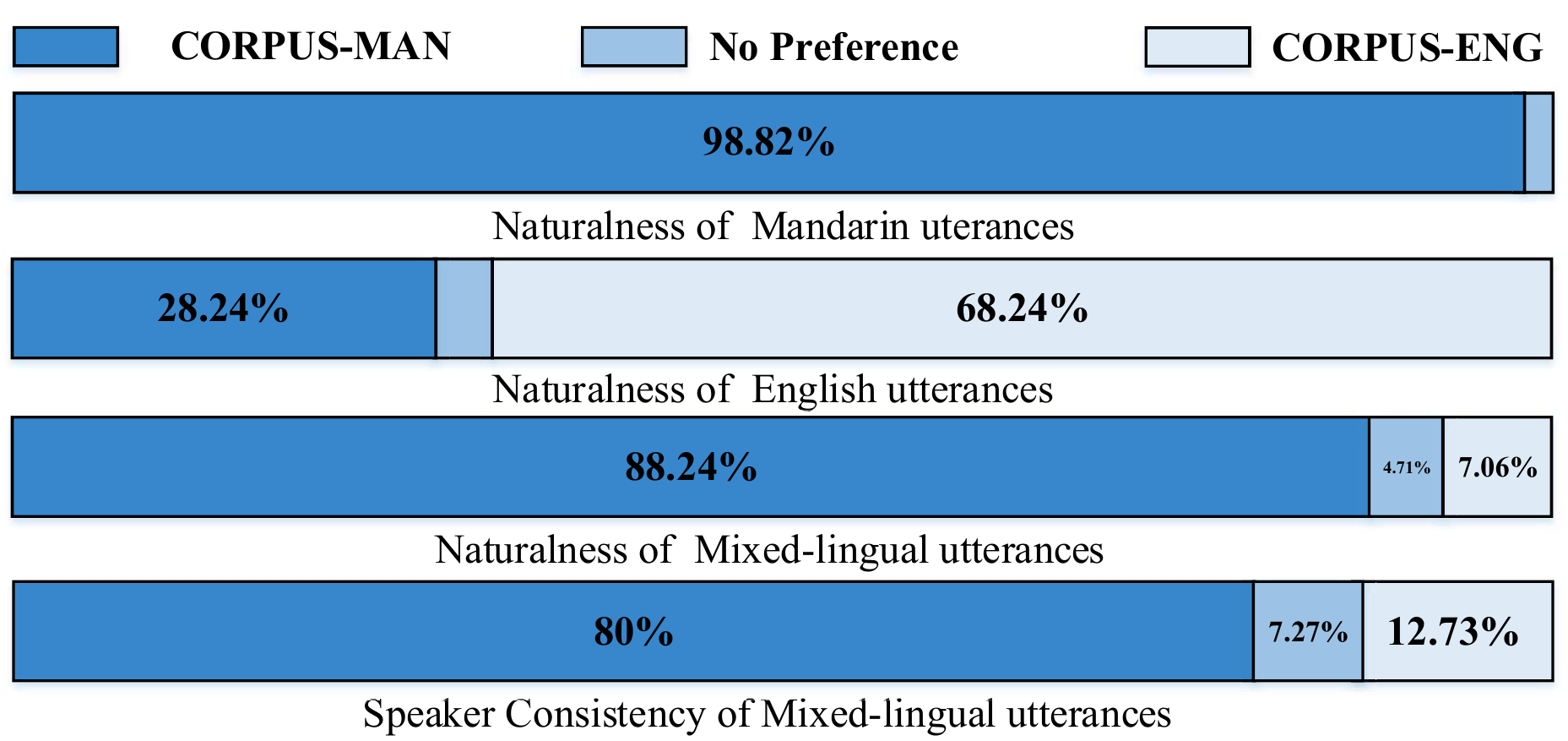}
   \caption{AB preference results of SE-DEC models trained using CORPUS-MAN and CORPUS-ENG.}
   \label{fig:cross_nat_smi_lang_cn_en}\vspace{-4pt}
 \end{figure}
\vspace{-4pt}

\subsection{Experiments analysis}\vspace{-2pt}
%\subsection{Different methods comparison}
We perform AB preference tests in terms of naturalness and speaker consistency to assess the performances of different methods. In detail, the speaker consistency AB preference tests are conducted on mixed-lingual sentences, which focus on the voice consistency within mixed sentences. Meanwhile, the naturalness AB preference tests are performed on Mandarin, English and mixed-lingual sentences. For each language, 30 sentences are randomly selected from test set. A group of 18 Mandarin listeners participates in the tests to give their preference choice\footnote{Samples can be found at \url{https://angelkeepmoving.github.io/mixed-lingual-tts/index.html} }.

\vspace{-8pt}
\subsubsection{Analysis of speaker embedding}\vspace{-3pt}
We first analyze the effect of speaker embedding by comparing different positions of speaker embedding. Here, the 500 Mandarin data of target speaker is mixed with data from the speakers for AVM. We compare the effects of speaker embedding at encoder output (SE-ENC) and decoder input (SE-DEC). The results show that SE-DEC brings better performances of speaker similarity and naturalness, as indicated in Figure~\ref{fig:ab_test_lang_enc_dec}. Placing speaker embedding at decoder input is surprisingly effective for mixed-lingual utterances. We argue that because speaker characteristic is more expressed in speech rather than text and the decoder is more related to speech in TTS task, placing the speaker embedding at the decoder input is more suitable.

%We first analyze the effect of speaker embedding by comparing different positions of speaker embedding in the encoder-decoder architecture. Here, the 500 Mandarin data of target speaker is mixed with data from the speakers for AVM. We compare the effects of speaker embedding at encoder output (SE-ENC) and decoder input (SE-DEC). Note that when speaker embedding is located at decoder input, it also affects the prenet, not only top layers in the decoder. The results show that SE-DEC brings better performances of speaker similarity and naturalness, as described in Figure~\ref{fig:ab_test_lang_enc_dec}. Placing speaker embedding at decoder input is surprisingly effective for mixed-lingual utterances. We argue that because the speaker embedding can affect the prenet when it is used at decoder input, it can handle language switching better and make the switch across languages more smooth.

%We then analyze how to put the speaker embedding for better voice consistency. Target and multi-speaker model with speaker embedding at encoder input (MTM-SE-ENC) and decoder output (MTM-SE-DEC) are compared. 

%\subsubsection{Retraining AVM versus Retraining speaker embedding}
\vspace{-8pt}
\subsubsection{Including versus excluding the target speaker data in the AVM training}\vspace{-3pt}
%{\color{red}IMPORTANT: Liumeng, we should have results to compare retraining AVM with MTM-SE-DEC. Although tomorrow is deadline, let's try to get this done.Otherwise, the logic is not smooth}
We then analyze how to use the data of the target speaker to generate mixed-lingual speech properly. The AVM excluding the target speaker data is pre-trained and used to adapt to the target speaker. We retrain decoder based on the AVM (Retrain-AVM) using 500 Mandarin data of the target speaker. Besides, as suggested by the results above, SE-DEC can achieve better performance than SE-ENC. Thus, the AVM including the target speaker data with speaker embedding at decoder (SE-DEC) is built to get the target speaker voice. We compare the performance between Retrain-AVM and SE-DEC. The results in Figure~\ref{fig:ab_test_lang_retrain_dec} show that the SE-DEC can achieve better performances than Retrain-AVM in terms of naturalness of three languages and speaker consistency. It suggests that including the target speaker data in AVM is helpful to build the mixed-lingual TTS system with monolingual data.

\vspace{-8pt}
\subsubsection{The choice of training data}
\begin{figure}[t]

   \centering
   \includegraphics[scale=0.38]{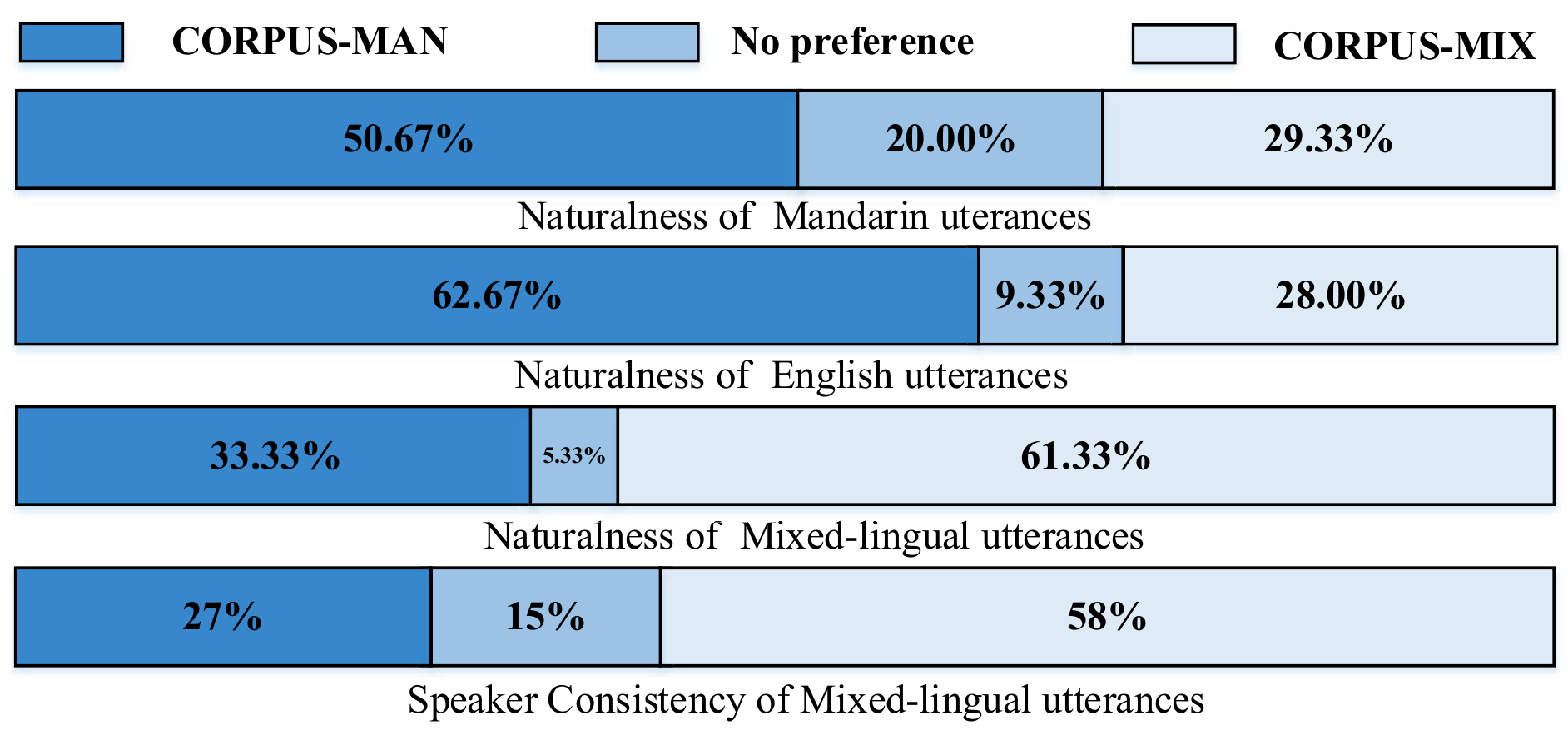}
   \caption{AB preference results of SE-DEC models trained using CORPUS-MAN and CORPUS-MIX.}
   \label{fig:cross_nat_smi_lang_mix_cn}\vspace{-6pt}
 \end{figure}

\begin{figure}[t]

   \centering
   \includegraphics[scale=0.38]{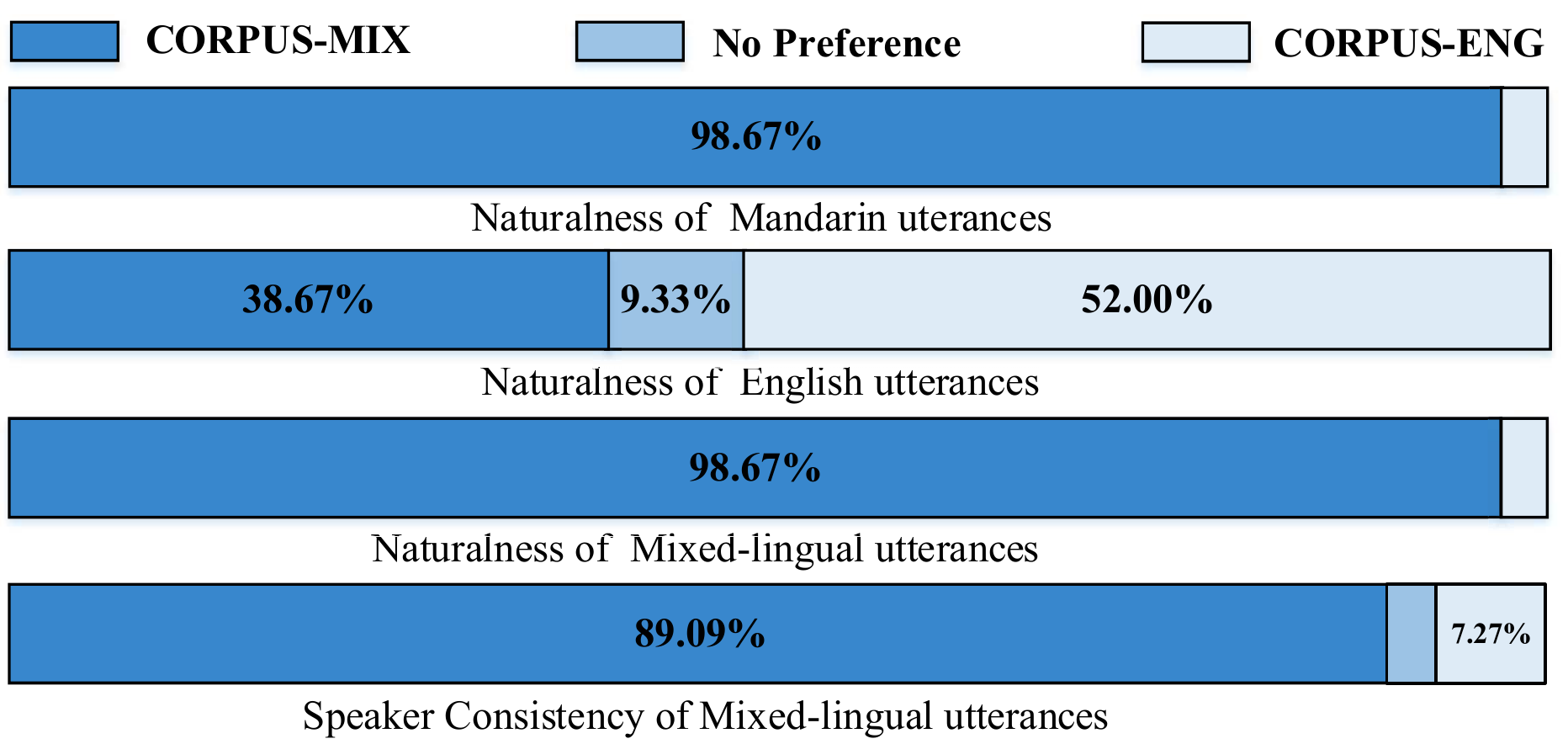}
   \caption{AB preference results of SE-DEC models trained using CORPUS-MIX and CORPUS-ENG.}
   \label{fig:cross_nat_smi_lang_mix_en}\vspace{-6pt}
 \end{figure}

\vspace{-3pt}
We then use the SE-DEC structure and study the impact of the choice of training data. Note that the data of target speaker is involved in the AVM training. We use three independent training sets from the same target speaker, 500 Mandarin utterances (CORPUS-MAN), 500 English utterances (CORPUS-ENG) and 500 mixed Mandarin-English utterances (CORPUS-MIX). This is to answer whether monolingual training data can achieve the same performance as that with mixed-lingual training data. The listening test results are presented in Figures~\ref{fig:cross_nat_smi_lang_cn_en},~\ref{fig:cross_nat_smi_lang_mix_cn}, and ~\ref{fig:cross_nat_smi_lang_mix_en}.  On the mixed-lingual test set, it is always preferred to have mixed-lingual training data. In the case of no mixed-lingual training data, listeners prefer the synthesized audio generated from model trained by Mandarin data. We argue that it may because the primary language of the built TTS system is Mandarin, and better Mandarin synthesis helps in the listening tests. To synthesize monolingual Mandarin, Mandarin training data is always preferred, followed by mixed-lingual. However, to synthesize monolingual English, even though English training data is always preferred, surprisingly Mandarin data is preferred than mixed-lingual data when no English training data is available. We plan to investigate on this aspect further.

 \begin{figure}[t]

   \centering
   \includegraphics[scale=0.38]{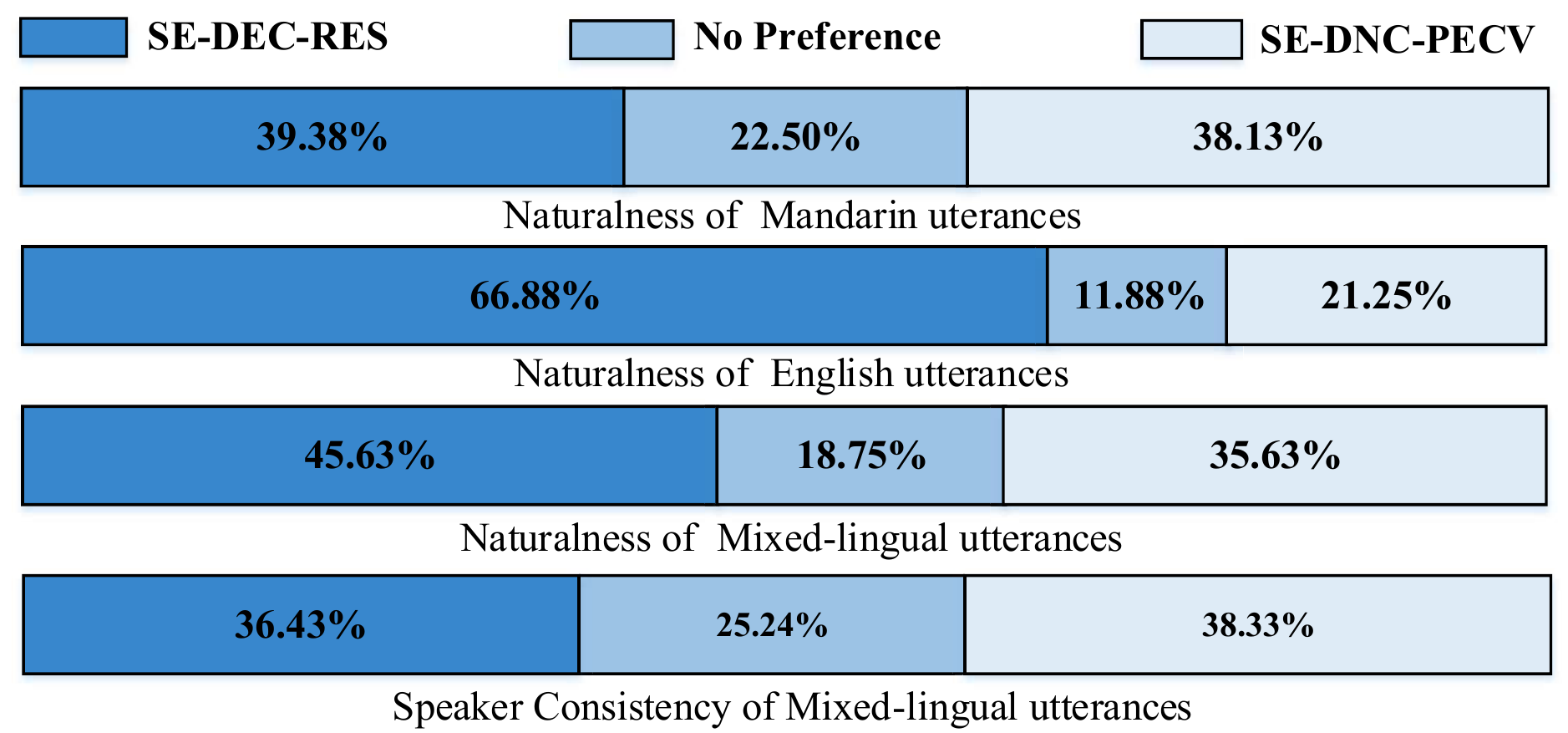}
   \caption{AB preference results of SE-DEC-RES and SE-DEC-PECV.}
   \label{fig:ab_test_lang_dec_c_r}\vspace{-6pt}
 \end{figure}

 \begin{figure}[t]

   \centering
   \includegraphics[scale=0.38]{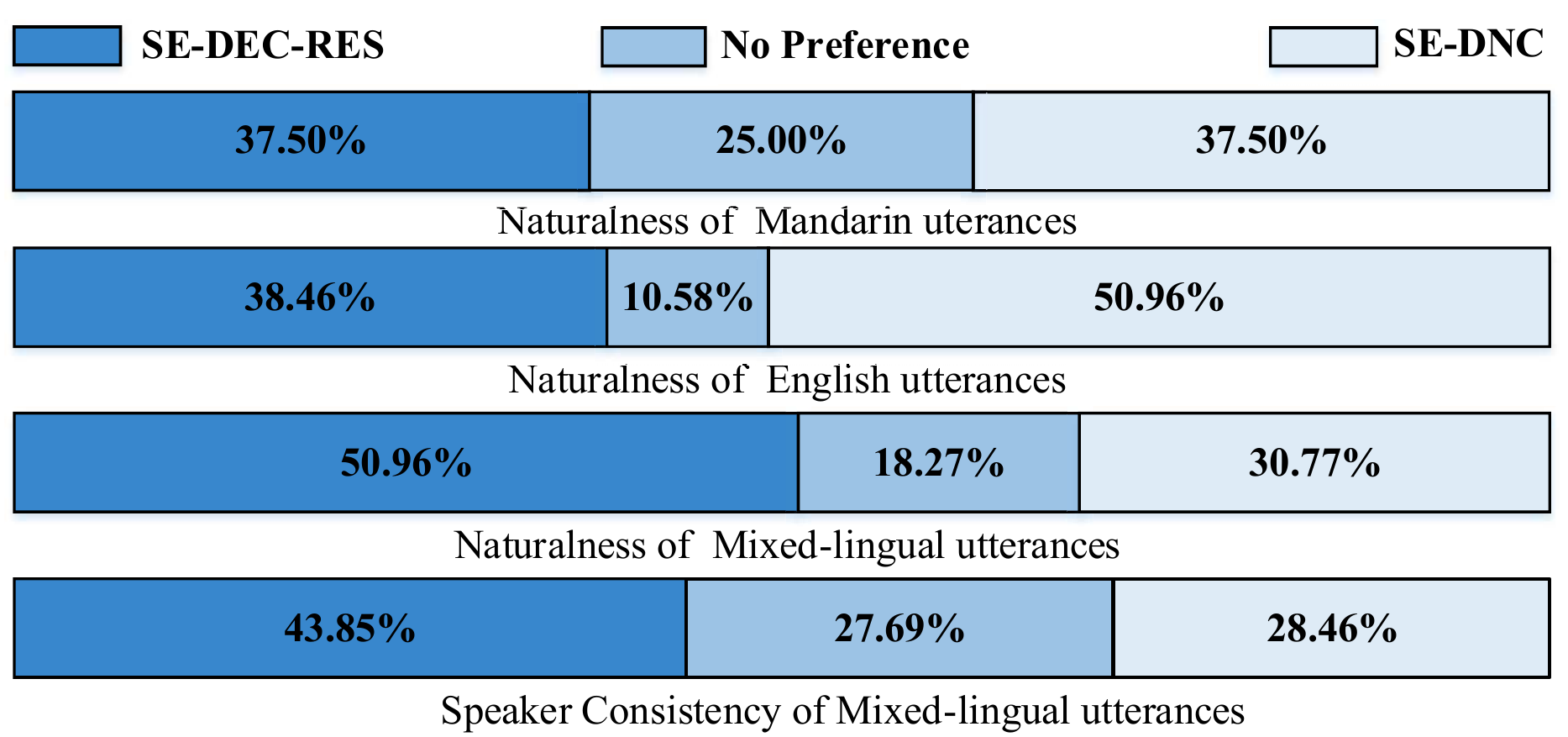}
   \caption{AB preference results of SE-DEC and SE-DEC-RES.}\vspace{-6pt}
   \label{fig:ab_test_lang_dec_r}
 \end{figure}
 
\vspace{-5pt}
\subsubsection{The use of phoneme-informed attention}\vspace{-4.1pt}
We also examine the impact of phoneme-informed attention. On the basis of SE-DEC model, two methods are performed: an additional phoneme embedding context vector (SE-DEC-PECV) and a residual encoder (SE-DEC-RES). Preference comparisons are presented in Figure~\ref{fig:ab_test_lang_dec_c_r}. The results demonstrate that using the residual encoder can achieve better naturalness than the additional context vector for three languages. For speaker consistency, the residual encoder and the additional context vector achieve almost the same preference. Furthermore, we compare the performance of using residual encoder or not. The results shown in Figure~\ref{fig:ab_test_lang_dec_r}, demonstrate that using residual encoder brings better naturalness for three languages, and also achieves better speaker consistency in mixed-lingual speech. 
%From the informal listening test, we find that phoneme-informed attention is useful to correct pronunciation errors, and we will investigate this in the future.

\vspace{-4pt}
\section{Conclusions}
In this paper, we investigate how to build a robust Mandarin-English mixed-lingual TTS system when only monolingual data of the target speaker is available. We conduct the study in the encoder-decoder framework. Here are our findings:
\begin{itemize}
\item The average voice model built from multi-speaker monolingual data is helpful, and the encoder part can learn phonetic information and the relationship between Mandarin and English phonemes. It also suggests that including the target speaker data in AVM training helps.
%\item {\color{blue}Average Voice Model built from multi-speaker monolingual data is helpful, and the AVM can learn phonetic relationship between Mandarin and English phonemes. It also suggests that including the target speaker data in AVM training helps.}
\item Speaker embedding is useful to control speaker identity and maintain speaker consistency within mixed-lingual utterances. Furthermore, the way to integrate speaker embedding in the encoder-decoder framework is important. Our experimental results show that integrating speaker embedding in the decoder input works better.
\item Although monolingual data is able to build a mixed-lingual TTS system, mixed-lingual training data is still preferred to have. It suggests that further investigations should be performed to built a better mixed-lingual TTS system with only monolingual data.
%It suggests to have further investigations to built a better mixed-lingual TTS system with only monolingual data.
\item Experimental results confirm that phoneme-informed attention not only helps with naturalness but also speaker consistency in mixed-lingual utterances.
\end{itemize}

From informal listening test, we find that  when only monolingual data is used to train the model, in some mixed-lingual utterances, the prosody is unnatural, specifically for Mandarin words next to English words, the tones of Mandarin words become inaccurate. In the future work, we plan to integrate our approach with a neural vocoder to produce better audio quality, and have a systematic investigation on phoneme embedding and speaker embedding, and also explore model training with more data.

%\\vspace{-4pt}
%\\section{Acknowledgements}
%\{\color{blue}\uwave{The research work is supported by the National Key Research and Development Program of China (No.2017YFB1002102).}}

%\newpage
%\balance
\bibliographystyle{IEEEtran}

\bibliography{template}

\begin{thebibliography}{10}
\providecommand{\url}[1]{#1}
\csname url@samestyle\endcsname
\providecommand{\newblock}{\relax}
\providecommand{\bibinfo}[2]{#2}
\providecommand{\BIBentrySTDinterwordspacing}{\spaceskip=0pt\relax}
\providecommand{\BIBentryALTinterwordstretchfactor}{4}
\providecommand{\BIBentryALTinterwordspacing}{\spaceskip=\fontdimen2\font plus
\BIBentryALTinterwordstretchfactor\fontdimen3\font minus
  \fontdimen4\font\relax}
\providecommand{\BIBforeignlanguage}[2]{{%
\expandafter\ifx\csname l@#1\endcsname\relax
\typeout{** WARNING: IEEEtran.bst: No hyphenation pattern has been}%
\typeout{** loaded for the language `#1'. Using the pattern for}%
\typeout{** the default language instead.}%
\else
\language=\csname l@#1\endcsname
\fi
#2}}
\providecommand{\BIBdecl}{\relax}
\BIBdecl

\bibitem{traber1999multilingual}
C.~Traber, K.~Huber, K.~Nedir, B.~Pfister, E.~Keller, and B.~Zellner, ``From
  multilingual to polyglot speech synthesis,'' in \emph{Sixth European
  Conference on Speech Communication and Technology}, 1999.

\bibitem{latorre2005polyglot}
J.~Latorre, K.~Iwano, and S.~Furui, ``Polyglot synthesis using a mixture of
  monolingual corpora,'' in \emph{Proceedings.(ICASSP'05). IEEE International
  Conference on Acoustics, Speech, and Signal Processing, 2005.}, vol.~1.\hskip
  1em plus 0.5em minus 0.4em\relax IEEE, 2005, pp. I--1.

\bibitem{liang2007hmm}
H.~Liang, Y.~Qian, and F.~K. Soong, ``An {HMM}-based bilingual
  ({M}andarin-{E}nglish) {TTS},'' \emph{Proceedings of SSW6}, 2007.

\bibitem{qian2008hmm}
Y.~Qian, H.~Cao, and F.~K. Soong, ``{HMM}-based mixed-language
  ({M}andarin-{E}nglish) speech synthesis,'' in \emph{2008 6th International
  Symposium on Chinese Spoken Language Processing}.\hskip 1em plus 0.5em minus
  0.4em\relax IEEE, 2008, pp. 1--4.

\bibitem{qian2009cross}
Y.~Qian, H.~Liang, and F.~K. Soong, ``A cross-language state sharing and
  mapping approach to bilingual ({M}andarin--{E}nglish) {TTS},'' \emph{IEEE
  Transactions on Audio, Speech, and Language Processing}, vol.~17, no.~6, pp.
  1231--1239, 2009.

\bibitem{he2012turning}
J.~He, Y.~Qian, F.~K. Soong, and S.~Zhao, ``Turning a monolingual speaker into
  multilingual for a mixed-language {TTS},'' in \emph{Thirteenth Annual
  Conference of the International Speech Communication Association}, 2012.

\bibitem{ramani2013voice}
B.~Ramani, M.~A. Jeeva, P.~Vijayalakshmi, and T.~Nagarajan, ``Voice
  conversion-based multilingual to polyglot speech synthesizer for {I}ndian
  languages,'' in \emph{2013 IEEE International Conference of IEEE Region 10
  (TENCON 2013)}.\hskip 1em plus 0.5em minus 0.4em\relax IEEE, 2013, pp. 1--4.

\bibitem{sitaram2016speech}
S.~Sitaram and A.~W. Black, ``Speech synthesis of code-mixed text.'' in
  \emph{LREC}, 2016.

\bibitem{sitaram2016experiments}
S.~Sitaram, S.~K. Rallabandi, S.~Rijhwani, and A.~W. Black, ``Experiments with
  cross-lingual systems for synthesis of code-mixed text.'' in \emph{SSW},
  2016, pp. 76--81.

\bibitem{chandu2017speech}
K.~R. Chandu, S.~K. Rallabandi, S.~Sitaram, and A.~W. Black, ``Speech synthesis
  for mixed-language navigation instructions.'' in \emph{INTERSPEECH}, 2017,
  pp. 57--61.

\bibitem{xie2016kl}
F.-L. Xie, F.~K. Soong, and H.~Li, ``A {KL} divergence and {DNN} approach to
  cross-lingual {TTS},'' in \emph{2016 IEEE International Conference on
  Acoustics, Speech and Signal Processing (ICASSP)}.\hskip 1em plus 0.5em minus
  0.4em\relax IEEE, 2016, pp. 5515--5519.

\bibitem{li2018bytes}
B.~Li, Y.~Zhang, T.~Sainath, Y.~Wu, and W.~Chan, ``Bytes are all you need:
  End-to-end multilingual speech recognition and synthesis with bytes,''
  \emph{arXiv preprint arXiv:1811.09021}, 2018.

\bibitem{Arik2018Neural}
S.~Arik, J.~Chen, K.~Peng, W.~Ping, and Y.~Zhou, ``Neural voice cloning with a
  few samples,'' in \emph{Advances in Neural Information Processing Systems},
  2018, pp. 10\,019--10\,029.

\bibitem{Nachmani2018Fitting}
E.~Nachmani, A.~Polyak, Y.~Taigman, and L.~Wolf, ``Fitting new speakers based
  on a short untranscribed sample,'' \emph{arXiv preprint arXiv:1802.06984},
  2018.

\bibitem{Ye2018Transfer}
Y.~Jia, Y.~Zhang, R.~Weiss, Q.~Wang, J.~Shen, F.~Ren, P.~Nguyen, R.~Pang, I.~L.
  Moreno, Y.~Wu \emph{et~al.}, ``Transfer learning from speaker verification to
  multispeaker text-to-speech synthesis,'' in \emph{Advances in Neural
  Information Processing Systems}, 2018, pp. 4480--4490.

\bibitem{Taigman2017VoiceLoop}
Y.~Taigman, L.~Wolf, A.~Polyak, and E.~Nachmani, ``Voiceloop: Voice fitting and
  synthesis via a phonological loop,'' \emph{arXiv preprint arXiv:1707.06588},
  2017.

\bibitem{gibiansky2017deep}
A.~Gibiansky, S.~Arik, G.~Diamos, J.~Miller, K.~Peng, W.~Ping, J.~Raiman, and
  Y.~Zhou, ``Deep voice 2: Multi-speaker neural text-to-speech,'' in
  \emph{Advances in neural information processing systems}, 2017, pp.
  2962--2970.

\bibitem{ping2018deep}
W.~Ping, K.~Peng, A.~Gibiansky, S.~O. Arik, A.~Kannan, S.~Narang, J.~Raiman,
  and J.~Miller, ``Deep voice 3: 2000-speaker neural text-to-speech,''
  \emph{Proc. ICLR}, 2018.

\bibitem{li2017deep}
C.~Li, X.~Ma, B.~Jiang, X.~Li, X.~Zhang, X.~Liu, Y.~Cao, A.~Kannan, and Z.~Zhu,
  ``Deep speaker: an end-to-end neural speaker embedding system,'' \emph{arXiv
  preprint arXiv:1705.02304}, 2017.

\bibitem{Skerry2018Towards}
R.~J. Skerry-Ryan, E.~Battenberg, X.~Ying, Y.~Wang, and R.~A. Saurous,
  ``Towards end-to-end prosody transfer for expressive speech synthesis with
  tacotron,'' 2018.

\bibitem{maaten2008visualizing}
L.~v.~d. Maaten and G.~Hinton, ``Visualizing data using t-{SNE},''
  \emph{Journal of machine learning research}, vol.~9, no. Nov, pp. 2579--2605,
  2008.

\bibitem{bahdanau2014neural}
D.~Bahdanau, K.~Cho, and Y.~Bengio, ``Neural machine translation by jointly
  learning to align and translate,'' \emph{arXiv preprint arXiv:1409.0473},
  2014.

\bibitem{veaux2016superseded}
C.~Veaux, J.~Yamagishi, K.~MacDonald \emph{et~al.}, ``Superseded-{CSTR} {VCTK}
  corpus: English multi-speaker corpus for {CSTR} voice cloning toolkit,''
  2016.

\bibitem{shen2018natural}
J.~Shen, R.~Pang, R.~J. Weiss, M.~Schuster, N.~Jaitly, Z.~Yang, Z.~Chen,
  Y.~Zhang, Y.~Wang, R.~Skerrv-Ryan \emph{et~al.}, ``Natural {TTS} synthesis by
  conditioning wavenet on mel spectrogram predictions,'' in \emph{2018 IEEE
  International Conference on Acoustics, Speech and Signal Processing
  (ICASSP)}.\hskip 1em plus 0.5em minus 0.4em\relax IEEE, 2018, pp. 4779--4783.

\bibitem{griffin1984signal}
D.~Griffin and J.~Lim, ``Signal estimation from modified short-time fourier
  transform,'' \emph{IEEE Transactions on Acoustics, Speech, and Signal
  Processing}, vol.~32, no.~2, pp. 236--243, 1984.

\bibitem{wang2017tacotron}
Y.~Wang, R.~Skerry-Ryan, D.~Stanton, Y.~Wu, R.~J. Weiss, N.~Jaitly, Z.~Yang,
  Y.~Xiao, Z.~Chen, S.~Bengio \emph{et~al.}, ``Tacotron: Towards end-to-end
  speech synthesis,'' \emph{arXiv preprint arXiv:1703.10135}, 2017.

\end{thebibliography}

% \begin{thebibliography}{9}
% \bibitem[1]{Davis80-COP}
%   S.\ B.\ Davis and P.\ Mermelstein,
%   ``Comparison of parametric representation for monosyllabic word recognition in continuously spoken sentences,''
%   \textit{IEEE Transactions on Acoustics, Speech and Signal Processing}, vol.~28, no.~4, pp.~357--366, 1980.
% \bibitem[2]{Rabiner89-ATO}
%   L.\ R.\ Rabiner,
%   ``A tutorial on hidden Markov models and selected applications in speech recognition,''
%   \textit{Proceedings of the IEEE}, vol.~77, no.~2, pp.~257-286, 1989.
% \bibitem[3]{Hastie09-TEO}
%   T.\ Hastie, R.\ Tibshirani, and J.\ Friedman,
%   \textit{The Elements of Statistical Learning -- Data Mining, Inference, and Prediction}.
%   New York: Springer, 2009.
% \bibitem[4]{YourName17-XXX}
%   F.\ Lastname1, F.\ Lastname2, and F.\ Lastname3,
%   ``Title of your INTERSPEECH 2019 publication,''
%   in \textit{Interspeech 2019 -- 20\textsuperscript{th} Annual Conference of the International Speech Communication Association, September 15-19, Graz, Austria, Proceedings, Proceedings}, 2019, pp.~100--104.
% \end{thebibliography}

\end{document}